\pdfoutput=1

\documentclass[11pt]{article}

\usepackage[final]{acl}

\usepackage{times}
\usepackage{latexsym}

\usepackage[T1]{fontenc}

\usepackage[utf8]{inputenc}

\usepackage{microtype}

\usepackage{inconsolata}

\usepackage{color}        
\usepackage{array}        
\usepackage{multirow}     
\usepackage{graphicx}     
\usepackage{float}        
\usepackage{bm}           
\usepackage{CJKutf8}      
\usepackage{amsmath}      
\usepackage{cleveref}     
\crefname{section}{§}{§§}
\Crefname{section}{§}{§§}
\usepackage{tikz}
\usepackage{natbib}
\usepackage{hyperref}
\usepackage{cleveref}

\title{Bi-DCSpell: A Bi-directional Detector-Corrector Interactive Framework for Chinese Spelling Check}

\author{Haiming Wu$^{1}$, Hanqing Zhang$^{1}$, Richeng Xuan$^{2}$ \and Dawei Song$^{1}$\thanks{Corresponding authors} \\
        $^{1}$School of Computer Science \& Technology, Beijing Institute of Technology, Beijing, China \\
        $^{2}$Beijing Academy of Artificial Intelligence (BAAI), Beijing, China \\
        \texttt{\{wuhm, zhanghanqing, dwsong\}@bit.edu.cn}, \texttt{rcxuan@baai.ac.cn}}

\begin{document}

\setlength{\lineskiplimit}{0pt}
\setlength{\lineskip}{0pt}
\setlength{\abovedisplayskip}{6pt}   
\setlength{\belowdisplayskip}{6pt}
\setlength{\abovedisplayshortskip}{6pt}
\setlength{\belowdisplayshortskip}{6pt} 

\maketitle
\begin{abstract}
Chinese Spelling Check (CSC) aims to detect and correct potentially misspelled characters in Chinese sentences.
Naturally, it involves the detection and correction subtasks, which interact with each other dynamically. Such interactions are bi-directional, i.e., the detection result would help reduce the risk of over-correction and under-correction while the knowledge learnt from correction would help prevent false detection.
Current CSC approaches are of two types: correction-only or single-directional detection-to-correction interactive frameworks. Nonetheless, they overlook the bi-directional interactions between detection and correction. This paper aims to fill the gap by proposing a \textbf{Bi}-directional \textbf{D}etector-\textbf{C}orrector framework for CSC (\textit{Bi-DCSpell}). Notably, \textit{Bi-DCSpell} contains separate detection and correction encoders, followed by a novel interactive learning module facilitating bi-directional feature interactions between detection and correction to improve each other's representation learning. Extensive experimental results demonstrate a robust correction performance of \textit{Bi-DCSpell} on widely used benchmarking datasets while possessing a satisfactory detection ability~\footnote{The source code will be made publicly available upon acceptance of the paper.}. 
\end{abstract}

\section{Introduction}
\label{chp1}
%

%
%
Chinese Spelling Check (CSC) aims to automatically detect and correct spelling errors in Chinese sentences \citep{yu2014chinese,wang2018hybrid,huang2021phmospell}. 
It serves as a foundation for a wide range of downstream applications in information retrieval (IR) and natural language processing (NLP), including Search Query Correction, 
Optical Character Recognition (OCR),
and Essay Scoring.
In recent years, CSC has garnered increasing attention from both academia and industry, emerging as a crucial task in the realm of Artificial Intelligence~\citep{zhang2020spelling,li2024mcrspell}.

\begin{figure}[tbp]
    \centering
    \includegraphics[width=0.4\textwidth]{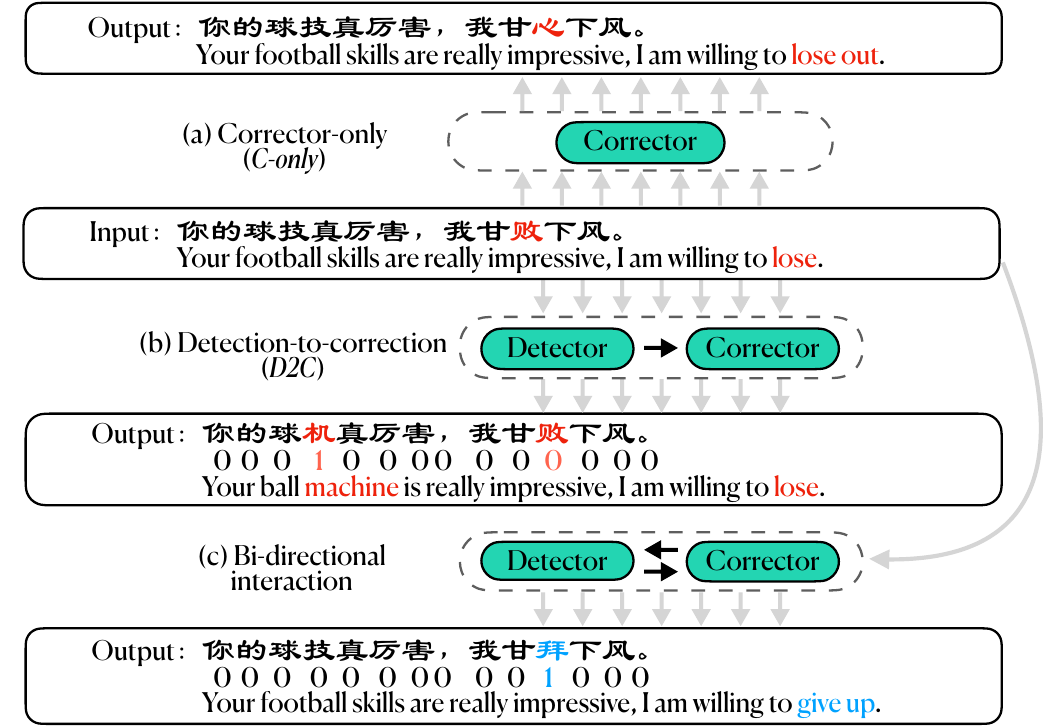}
    \caption{Comparison of three types of interactions: `correction-only', `detection-to-correction' and `bi-directional interactions'. The \textcolor{red}{wrongly}/\textcolor{blue}{ground-truth} corrected characters in the candidates are in \textcolor{red}{red}/\textcolor{blue}{blue}.}
    \label{fig.example}
\end{figure}

Most existing approaches to CSC can be categorized into two types of framework: (a) correction-only (\textit{C-only}) and (b) detection-to-correction (\textit{D2C}). The \textit{C-only} does not contain a detection module. Instead, it directly adopts a specially designed corrector to perform the correction task \cite{hong2019faspell, zhang2020spelling, cheng2020spellgcn, li2021dcspell, guo-etal-2021-global}. The most recent \textit{C-only} models are inspired by masked language model (MLM) \cite{devlin2019bert}, where each character to be corrected is predicted using contextual information. The second type of framework possesses both detection and correction modules but the interaction between them is single-directional in a way akin to a pipeline \cite{cheng2020spellgcn, zhang2021correcting, zhu2022mdcspell}. That is, the detector first identifies the positions of errors, which are then used as prior knowledge for the correction module. While these two types of framework have been the dominant paradigms in the current CSC literature, they neglect the bi-directional interactions between detector and corrector. This oversight leads to issues such as the substitution of correct characters with incorrect ones (\textit{over-correction}) or the failure to correct misspelled characters (\textit{under-correction}). 

Figure~\ref{fig.example} presents an example to illustrate the differences among \textit{C-only}, \textit{D2C}, and bi-directional interactions. The input ``\begin{CJK}{UTF8}{gbsn}你的球机真厉害，我甘败下风。\end{CJK}" contains two misspelled characters ``\begin{CJK}{UTF8}{gbsn}机(jī, machine)\end{CJK}" and ``\begin{CJK}{UTF8}{gbsn}败(bài, fail)\end{CJK}", which should be corrected as ``\begin{CJK}{UTF8}{gbsn}技(jì, skill)\end{CJK}" and ``\begin{CJK}{UTF8}{gbsn}拜(bài, worship)\end{CJK}", respectively. Figure 1(a) illustrates the \textit{C-only} approaches that are typically underpinned by an MLM. Note that MLM tends to rectify a correct low-frequency collocation into a high-frequency one \cite{yang2022cospa}. In this example, as ``\begin{CJK}{UTF8}{gbsn}败(bài, fail)\end{CJK}" is more frequent in natural language than ``\begin{CJK}{UTF8}{gbsn}拜(bài, worship)\end{CJK}", it is difficult for a \textit{C-only} approach to correct ``\begin{CJK}{UTF8}{gbsn}败(bài, fail)\end{CJK}" properly in case there is no prior error information from detection. 
In Figure 1(b), the \textit{D2C} interaction first uses an error detector that determines ``\begin{CJK}{UTF8}{gbsn}败(bài, fail)\end{CJK}" is wrong. The result is then fed into the correction module, which then successfully corrects ``\begin{CJK}{UTF8}{gbsn}败(bài, fail)\end{CJK}" as ``\begin{CJK}{UTF8}{gbsn}拜(bài, worship)\end{CJK}". However, the corrector still may not be able to correct the other misspelled character ``\begin{CJK}{UTF8}{gbsn}机(jī, machine)\end{CJK}" as it is not pre-detected by the detector. 
This reflects that one-directional interaction (from detection to correction) can still lead to under-correction or over-correction.

To fill this gap, we propose a novel \textbf{Bi}-directional \textbf{D}etector-\textbf{C}orrector interactive framework for \textbf{Spell}ing check (\textit{Bi-DCSpell}). The detection and correction subtasks dynamically interact with each other, implying that knowledge acquired from one can be transferred to enhance the representation learning of the other. This bi-directional exchange of knowledge facilitates mutual improvement in performance. A concrete case is depicted in Figure~\ref{fig.example}(c). Different from the \textit{C-only} and \textit{D2C} approaches, \textit{Bi-DCSpell} not only uses the detection result (``\begin{CJK}{UTF8}{gbsn}败(bài, fail)\end{CJK}" is wrong), for the corrector to revise ``\begin{CJK}{UTF8}{gbsn}败(bài, fail)\end{CJK}" as ``\begin{CJK}{UTF8}{gbsn}拜(bài, worship)\end{CJK}", but also feeds the correction result 
back to the detector as contextual information to help identify that  ``\begin{CJK}{UTF8}{gbsn}机(jī, machine)\end{CJK}" may also be wrong. 

Specifically, our approach uses two independent encoders: one dedicated to detection and the other to correction, to capture the subtask-specific features from the input sequence. Subsequently, we introduce an interactive learning module, which comprises a series of bi-directional cross-attention layers to enable the dynamic interactions between the two subtasks. 
Finally, two task-specific classifiers are used to output the detection labels and generate the correction sequence. The performance of our method is evaluated on three widely used CSC benchmark datasets: SIGHAN13, SIGHAN14, and SIGHAN15. The results show that \textit{Bi-DCSpell} surpasses interaction-free and uni-directional interactive baselines by 3.0\% and 1.8\% respectively in correction F1 score. Moreover, our method compares favorably against the current state-of-the-art approaches on all three datasets. A further ablation study demonstrates the effect of incorporating the bi-directional interactions in CSC. We also present a case study with concrete examples to illustrate the advantages of our model. 

\section{Related Work}
\label{chp2}
\subsection{Correction-only Approaches for CSC}
Correction-only (\textit{C-only}) is a kind of approaches in CSC, employing a corrector to directly correct the input sequence without error detection. Recent approaches along this line with a masked language model (MLM), and have achieved significant performance improvements
\cite{hong2019faspell, wang2019confusionset, zhang2020spelling, zhang2021correcting, huang2021phmospell, zhu2022mdcspell}.
For instance, 
SpellGCN \cite{cheng2020spellgcn} integrates phonological and visual similarity information into character classifiers using a graph network, which then feeds the graph representation into MLM.
ReaLise \cite{xu2021read} employs three distinct feature networks to capture phonetic, graphemic and semantic features, ultimately passing the fused representation through MLM.
These methods focus on constructing CSC features and feeding them into an MLM-based corrector. 
In contrast, FASPell \cite{hong2019faspell} leverages phonological and visual similarity features to construct a filtering model, selects the most suitable candidate Chinese characters from a pre-trained Language Model (PLM). To better formalize this type of CSC methods, we denote the model parameters as $\bm{\theta}_{c}$, and the correction inference degenerates to:
\begin{equation}
\begin{aligned}
    P(y^C_i|\bm{x})=P(y^C_i|\bm{x}, \bm{\theta}_{c})
\end{aligned}
\end{equation}
where, $y^C_i$ represents the corrected character at the $i$-th position in the input sequence $\bm{x}$, with $C$ denoting the correction sub-task identifier.


\subsection{Detection-to-correction for CSC}
Detection-to-correction (\textit{D2C}) represents the second type of existing CSC approaches, which first use an error detector to detect the position information of misspelled characters and then feed the detection results to the corrector.  
Most approaches employ a conventional multiple-stage pipeline approach, such as Soft-masked BERT \cite{zhang2020spelling}, MLM-phonetics \cite{zhang2021correcting} and DR-CSC \cite{huang2023frustratingly}.
%
Unlike these pipeline \textit{D2C} approaches, MDCSpell \cite{zhu2022mdcspell} uses parallel detection and correction feature representation modules, and the corrector receives the detector's hidden states, thus, the inference in correction incorporates the feature from both detection and correction. Accordingly, \textit{D2C} approaches can be formalized as:
\begin{equation}
\begin{aligned}
    P(y^D_i|\bm{x})&=P(y^D_i|\bm{x}, \bm{\theta}_{d})  \\
    P(y^C_i|\bm{x})&=P(y^C_i|\bm{x}, \bm{\theta}_{d}, \bm{\theta}_{c})
\end{aligned}
\end{equation}
where $\bm{\theta}_{d}$ denotes the parameters of detection.

Similar to MDCSpell, our method also employs parallel detection and correction processes. However, a key distinction is that, 
our model facilitates a bi-directional interactive exchange of information between the detection and correction processes, i.e., the features learned from one subtask are utilized as prior knowledge for the other. We expect such bi-directional interactions would reinforce the representation learning for both subtasks.

\section{Bi-directional Interactive Framework}
\label{chp3}
\begin{figure*}[t]
    \centering
    \includegraphics[width=1.0\textwidth]{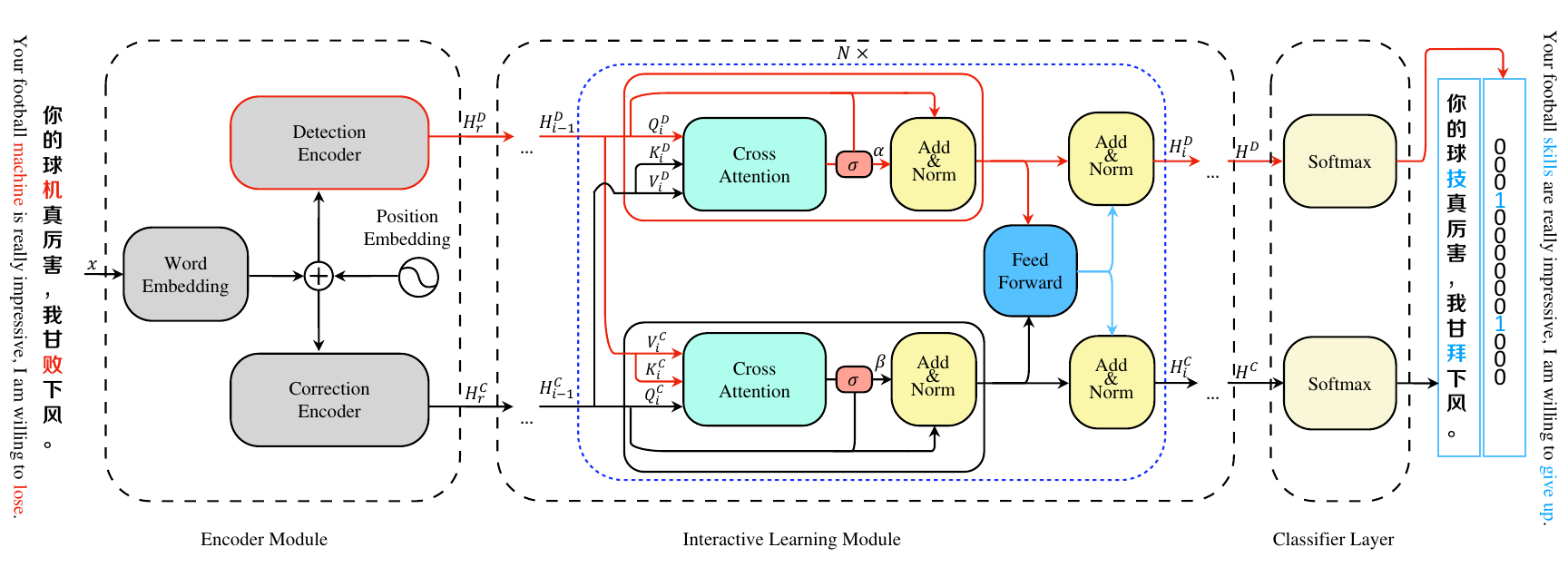}
    \caption{Overview of our \textit{Bi-DCSpell} framework. $\oplus$ is the matrix addition operation, and $\sigma$ denotes learnable control gate.}
    \label{fig.model}
\end{figure*}
\subsection{Problem Formulation}
\label{sec:problem}
The Chinese Spelling Check (CSC) task aims to detect and correct the misspelled characters in a Chinese sentence. It involves two sub-tasks: detection and correction. Thus it becomes a multi-task learning problem. Formally, given an input textual sequence $\bm{x} = (w_1, w_2, \cdots, w_n)$, where each Chinese character $w_i$ is taken from a vocabulary $\bm{V}$, a CSC model needs to detect whether each character is erroneous (i.e., misspelled), so as to output a sequence of detection labels $\bm{y}^D = (y^D_1, y^D_2, \cdots, y^D_n)$ and generate the corresponding corrections $\bm{y}^C = ( y^C_1, y^C_2, \cdots, y^C_n )$, where $y^D_i \in \{0,1\}$, $y^C_i \in \bm{V}$. 

\textbf{Bi-directional Interaction}
In this paper, we hypothesize that the bi-directional interaction between detection and correction would benefit the performance improvement of two sub-tasks, i.e., alleviating the challenges of under-correction and over-correction present in current approaches. To achieve this, we have crafted an interactive learning module that facilitates bi-directional transfer of hidden knowledge (in term of features) between detection and correction. As shown in Figure~\ref{fig.example}(c), when the system detects whether the character in input text is incorrect, the detection module integrates feature information from correction for the detection process. Simultaneously, the correction module incorporates detection features from the detection module to make corrections. Therefore, the incorporation of bi-directional interactions in CSC can be formalized as computing the following probability for $y_i^D, y_i^C$:
\begin{equation}
\begin{aligned}
    P(y^D_i, y^C_i|\bm{x})=P&(y_i^D, y^C_i|\bm{x}, \bm{\theta}_{d}, \bm{\theta}_{c})
\end{aligned}
\end{equation}

\subsection{\textit{Bi-DCSpell} for CSC}
We propose a \textit{Bi-directational Detector-Corrector} interactive framework for Chinese spelling check, namely \textit{Bi-DCSpell}. The framework comprises three components: an Encoder module, an Interactive Learning module, and a Classifier Layer. The overall structure is depicted in Figure~\ref{fig.model}.

\subsubsection{Detection and Correction Encoders}

To capture the subtask-specific features, two separate encoders are used, one for detection and the other for correction. The input text is first mapped to a sequence of embedding vectors by the embedding layer ${\rm Embedding}$. Then, \textit{Bi-DCSpell} takes two separate encoders for detection and correction (${\rm Encoder}^D$ and ${\rm Encoder}^C$), formalized as:
\begin{equation}
\begin{aligned}
    &\bm{X} = {\rm Embedding}(\bm{x})          \\
    &\bm{H}^D_r = {\rm Encoder}^D(\bm{X})      \\
    &\bm{H}^C_r = {\rm Encoder}^C(\bm{X}),
\end{aligned}
\end{equation}
where $\bm{H}^D_r, \bm{H}^C_r$ denote detection-specific and correction-specific feature representations respectively, which are essentially the hidden states without any cross-task interaction. To facilitate the calculation in the interactive learning module, we keep $\bm{H}^D_r$ and $\bm{H}^C_r$ at the same size $R^{n \times d_h}$, where $d_h$ is the dimension of hidden state.

\subsubsection{Interactive Learning}

To model the bi-directional interactions between detection and correction tasks, we design an interactive learning module as shown in the middle part of Figure~\ref{fig.model}. It consists of a series of stacked bi-directional interaction layers similar to the Transformer \cite{vaswani2017attention}, with each layer incorporating two task-specific cross-attention networks, two learnable control gates and a feed-forward network.



\textbf{Task-specific Cross-attention Network.} Firstly, two task-specific cross-attention networks use the feature knowledge learnt from one task to inform the representation learning for the other, respectively. Here, we use the same structure as vanilla attention \cite{vaswani2017attention} in each cross-attention process, but the input involves two representation states (i.e., detection and correction) instead of one.

In detection cross-attention, the detection representation is updated by extracting corresponding information in correction features as follows:
\begin{equation}
    \tilde{\bm{H}}^D_{is} = softmax (\frac{\bm{Q}_i^D {\bm{K}_i^D}^{\mathsf{T}}}{\sqrt{d_K}}) \bm{V}_i^D
\end{equation}
where, $\tilde{\bm{H}}^D_{is}$ denotes the hidden state of interactive detection, and $softmax$ is the $softmax$ function, and $\mathsf{T}$ denotes the transpose of matrix. The set of matrices (i.e., queries $\bm{Q}_i^D$, keys $\bm{K}_i^D$, and values $\bm{V}_i^D$) is calculated by linear transformations from $\bm{H}_{i-1}^D$ and $\bm{H}_{i-1}^{C}$. Specifically, $\bm{H}_{i-1}^{D}$ is used to compute queries $\bm{Q}_i^D$, and $\bm{H}_{i-1}^{C}$ is used to compute values $\bm{V}_i^D$ and keys $\bm{K}_i^D$, where all the parameters in the linear functions are trainable. 

And, similar to the calculation process of $\tilde{\bm{H}}_{is}^{D}$, we take $\bm{H}_{i-1}^C$ and $\bm{H}_{i-1}^{D}$ to compute queries $\bm{Q}_i^C$, keys $\bm{K}_i^C$ and values $\bm{V}_i^C$, and then utilize the three matrices ($\bm{Q}_i^C, \bm{K}_i^C, \bm{V}_i^C$) to update the correction representation by extracting the error position information in detection features $\bm{H}_{i-1}^{D}$. 
\begin{equation}
    \begin{aligned}
        \tilde{\bm{H}}^C_{is} &= softmax (\frac{\bm{Q}_i^C {\bm{K}_i^C}^{\mathsf{T}}}{\sqrt{d_K}}) \bm{V}_i^C
    \end{aligned}
\end{equation}
where $\tilde{\bm{H}}_{is}^{C}$ denotes the hidden state of interactive correction. 

\textbf{Learnable Control Gates} Then, considering that the degree of interaction between the correction and the detection is different for each example, we employ two learnable control gates ($\alpha$ and $\beta$) to regulate the extent of interaction between one sub-task and the other:
\begin{equation}
    \begin{aligned}
        \alpha &= \sigma (W_{i}^D [ \tilde{\bm{H}}_{is}^D, \bm{H}_{i-1}^D ]+b_{i}^D) \\
        \bm{H}_{is}^{D'} =& LN(\alpha \odot \tilde{\bm{H}}_{is}^D + (1-\alpha) \odot \bm{H}_{i-1}^D) \\
        \beta &= \sigma (W_{i}^C [ \tilde{\bm{H}}^C, \bm{H}_{i-1}^C ]+b_{i}^C)  \\
        \bm{H}_{is}^{C'} =& LN(\beta \odot \tilde{\bm{H}}_{is}^C + (1-\beta) \odot \bm{H}_{i-1}^C) 
    \end{aligned}
\end{equation}
where, $(W_{i}^*, b_{i}^*)$ are learnable gate parameters and $LN$ denotes the layer normalization. $\bm{H}_{is}^{D'}$ and $\bm{H}_{is}^{C'}$ represent detection and correction hidden states after incorporating the correction/detection features in detection/correction representation learning.

\textbf{Feed-forward Network.} To merge the detection and correction feature knowledge, we designed a new feed-forward network inspired by Transformer. First, we concatenate the detection and correction hidden states and then feed it into a linear multiplication layer to learn the merged features.
\begin{equation}
    \begin{aligned}
        \bm{H}_{i}^{DC} &= \bm{H}_{is}^{D'} \oplus \bm{H}_{is}^{C'}  \\
        \bm{H}_i^{DC'} = \bm{W}_2 & [{\rm ReLU}(\bm{W}_1 \bm{H}_i^{DC} + \bm{b}_1)] + \bm{b}_2
    \end{aligned}
\end{equation}
where $\oplus$ denotes the tensor concatenation. ${\rm ReLU}$ is the $relu$ activation function, and $(\bm{W}_1, \bm{b}_1)$, $(\bm{W}_2, \bm{b}_2)$ are two sets of parameters.

Finally, the merged hidden states $\bm{H}_i^{DC'}$ are projected into a normalization function, and added to $\bm{H}^{D'}_{is}$ and $\bm{H}^{C'}_{is}$ to update the detection and correction representations, respectively.
\begin{equation}
    \begin{aligned}
        \bm{H}_i^D = LN (\bm{H}^{D'}_{is} + {\bm{H}_i^{DC'}}) \\
        \bm{H}_i^C = LN (\bm{H}^{C'}_{is} + {\bm{H}_i^{DC'}})
    \end{aligned}
\end{equation}
where $\bm{H}_i^D$ and $\bm{H}_i^C$ are the updated feature representations. In the last bi-directional interaction layer $L$, the updated representations $\bm{H}_L^D$ and $\bm{H}_L^C$ are utilized as the ultimate states $\bm{H}^D$ and $\bm{H}^C$.

\subsubsection{Classifier Layer}
Hidden states $\bm{H}^D$ and $\bm{H}^C$ are fed into two task-specific classifiers to predict the detection label and generate the correction character for each corresponding character. 
Given the original input sentence $\bm{x}$ and its two representation ($\bm{H}^D$, $\bm{H}^C$), we can compute its probability distributions $p(\hat{y}_i^D|\bm{x})$ and $\hat{y}_i^C$ for $i$-th character.
\begin{equation}
\label{softmax}
    \begin{aligned}
        \hat{y}_i^D = {\rm softmax}(\bm{W}_D \bm{h}^D_i + \bm{b}_D)  \\
        \hat{y}_i^C = {\rm softmax}(\bm{W}_C \bm{h}^C_i + \bm{b}_C)
    \end{aligned}
\end{equation}
where ($\bm{W}_D, \bm{b}_D$) and ($\bm{W}_C, \bm{b}_C$) are trainable parameters. Thus, the $\hat{y}_i^D$ and $\hat{y}_i^C$ denote output distributions of detection and correction, respectively.

\subsubsection{Model Training}
Given a set of manually labeled training data $\{(\bm{x}_j, \bm{y}^D_j, \bm{y}^C_j)\}^m_{j=1}$ ($m$ denotes the number of training samples), we use cross-entropy loss as objective functions to train the above model. To unify the training process, a linear multi-objective optimization function is designed to guide the learning of the model to balance the detection and correction processes.
\begin{equation}
    \begin{aligned}
    \mathcal{L}^D =& -\sum\nolimits_{j=1}^{m} \sum\nolimits_{i=1}^{n} y^D_i \log (\hat{y}^D_i) \\
    \mathcal{L}^C =& -\sum\nolimits_{j=1}^{m} \sum\nolimits_{i=1}^{n} y^C_i \log (\hat{y}^C_i) \\
    \mathcal{L} &= \lambda \mathcal{L}^C + (1-\lambda)\mathcal{L}^D
    \end{aligned}
\end{equation}
where, the $\lambda \in [0,1]$ is a transfer-balanced hyperparameter to guide model learning.

\section{Experiments}
\label{chp4}
\subsection{Datasets and Pre-Processing}
Extensive empirical evaluation is carried out on three widely used CSC datasets: SIGHAN13 \cite{wu2013chinese}, SIGHAN14 \cite{yu2014overview} and SIGHAN15 \cite{tseng2015introduction}. We train \textit{Bi-DCSpell} using four datasets, three of which are training data sets from SIGHAN, with approximately 10,000 data samples. The fourth dataset is an additional set of training data generated by an automatic method \citep{wang2018hybrid} with 271,009 samples. 

Consistent with the current literature \cite{xu2021read, zhu2022mdcspell}, we use three test datasets from SIGHAN13, SIGHAN14 and SIGHAN15 to evaluate the trained model. Furthermore, we use the same pre-processing procedure and transform the characters in these datasets to simplified Chinese with OpenCC\footnote{https://github.com/BYVoid/OpenCC}. 

The evaluation metrics are provided in appendix~\ref{appendix_evaluation_metrics}.  

\subsection{Implementation details}

We use PyTorch \cite{paszke2019pytorch} to implement the proposed \textit{Bi-DCSpell} the Transformers library \cite{wolf2020transformers}.
For scientific comparison with existing methods, two pre-trained language models,  Chinese-BERT-wwm \cite{cui2021pre} (abbreviated as BERT) and ChineseBERT \cite{sun2021chinesebert}, are used to initialize the embedding layer and the correction encoder, each model has 12 transformer layers with 12 attention heads and outputs for each token a hidden representation of dimensionality 768. We initialize the weights of the detector encoder using the Transformer parameters of the bottom two layers in the pre-trained language model.
For model training,  AdamW \cite{loshchilov2018fixing} is used as an optimizer with max epochs 20, the learning rate is set as 5e-5, and the batch size is set to 32. The training process takes about 10 hours on a single RTX A6000 (48GB GPU memory).

\begin{table*}[ht]
\centering
\small
\renewcommand{\arraystretch}{1.0}  
    \begin{tabular}{c|llcccccc}
    \hline
    \multicolumn{1}{c|}{\multirow{2}{*}{\textbf{Dataset}}}   &\multicolumn{1}{c|}{\multirow{2}{*}{\textbf{Baseline}}}    &\multicolumn{1}{c|}{\multirow{2}{*}{\textbf{Backbone}}}      & \multicolumn{3}{c|}{\textbf{Detection}}   & \multicolumn{3}{c}{\textbf{Correction}}    \\ 
    \cline{4-9} 
    \multicolumn{1}{c|}{}   &\multicolumn{1}{c|}{}     &\multicolumn{1}{c|}{}      & \textbf{Prec.} & \textbf{Rec.} & \multicolumn{1}{c|}{\textbf{F1}}  & \textbf{Prec.} & \textbf{Rec.} & \textbf{F1} \\ 
    \hline
    \multirow{11}{*}{SIGHAN13}    
        &\multicolumn{1}{l|}{\emph{Soft-Masked BERT $^{\ast}$} \cite{zhang2020spelling}}   &\multicolumn{1}{l|}{BERT}      & 81.1     & 75.7     & \multicolumn{1}{c|}{78.3}     & 75.1     & 70.1     & 72.5     \\
        &\multicolumn{1}{l|}{\emph{SpellGCN} \cite{cheng2020spellgcn}}   &\multicolumn{1}{l|}{BERT}  & 80.1     & 74.4     & 
            \multicolumn{1}{c|}{77.2}     & 78.3     & 72.7     & 75.4     \\
        &\multicolumn{1}{l|}{\emph{MLM-phonetics} \cite{huang2021phmospell}}   &\multicolumn{1}{l|}{BERT}  & 82.0      & 78.3     & \multicolumn{1}{c|}{80.1}     & 79.5     & 77.0     & 78.2     \\ 
        &\multicolumn{1}{l|}{\emph{DCN} \cite{wang-etal-2021-dynamic}}   &\multicolumn{1}{l|}{BERT}  & 86.8     & 79.6     & \multicolumn{1}{c|}{83.0}     & 84.7     & 77.7     & 81.0     \\
        &\multicolumn{1}{l|}{\emph{GAD} \cite{guo-etal-2021-global}}    &\multicolumn{1}{l|}{BERT}   & 85.7    & 79.5    & \multicolumn{1}{c|}{82.5}     & 84.9    & 78.7    & 81.6    \\
        &\multicolumn{1}{l|}{\emph{MDCSpell} \cite{zhu2022mdcspell}}    &\multicolumn{1}{l|}{BERT}   & \textbf{89.2}      & 78.3     & \multicolumn{1}{c|}{83.4}     & 87.5     & 76.8     & 81.8     \\  
        &\multicolumn{1}{l|}{\emph{DORM} \cite{liang2023disentangled}}    &\multicolumn{1}{l|}{ChineseBERT}   & 87.9      & 83.7     & \multicolumn{1}{c|}{85.8}     & 86.8     & 82.7     & 84.7     \\  
        &\multicolumn{1}{l|}{\emph{DR-CSC} \cite{huang2023frustratingly}}     &\multicolumn{1}{l|}{ChineseBERT} & 88.5      & 83.7     & \multicolumn{1}{c|}{86.0}     & 87.7     & 83.0     & 85.3     \\  
        \cline{2-9}
        &\multicolumn{1}{l|}{PLM-FT (ChineseBERT)}     &\multicolumn{1}{l|}{ChineseBERT}   & 86.9     & 82.0      & \multicolumn{1}{c|}{84.4}        & 85.6    & 80.7     & 83.1         \\ 
        &\multicolumn{1}{l|}{\textit{Bi-DCSpell} (BERT)}     &\multicolumn{1}{l|}{BERT}   & 88.2     & 80.6      & \multicolumn{1}{c|}{84.2}       & 86.8     & 78.7      & 82.6       \\ 
        &\multicolumn{1}{l|}{\textit{Bi-DCSpell} (ChineseBERT)}     &\multicolumn{1}{l|}{ChineseBERT}   & 89.0     & \textbf{85.1}      & \multicolumn{1}{c|}{\textbf{87.0}}       & \textbf{87.8}     & \textbf{84.1}    & \textbf{85.9}       \\ 
        \hline
    \multirow{11}{*}{SIGHAN14}    
        &\multicolumn{1}{l|}{\emph{Soft-Masked BERT $^{\ast}$} \cite{zhang2020spelling}}    &\multicolumn{1}{l|}{BERT}    & 65.2     & 70.4     & \multicolumn{1}{c|}{67.7}     & 63.7     & 68.7     & 66.1     \\
        &\multicolumn{1}{l|}{\emph{SpellGCN} \cite{cheng2020spellgcn}}    &\multicolumn{1}{l|}{BERT}    & 65.1    & 69.5    & \multicolumn{1}{c|}{67.2}     & 63.1    & 67.2    & 65.3    \\
        &\multicolumn{1}{l|}{\emph{MLM-phonetics} \cite{zhang2021correcting}}     &\multicolumn{1}{l|}{BERT}    & 66.2     & \textbf{73.8}     & \multicolumn{1}{c|}{69.8}     & 64.2     & \textbf{73.8}     & 68.7     \\ 
        &\multicolumn{1}{l|}{\emph{DCN} \cite{wang-etal-2021-dynamic}}     &\multicolumn{1}{l|}{BERT}    & 67.4     & 70.4     & \multicolumn{1}{c|}{68.9}     & 65.8     & 68.7     & 67.2     \\
        &\multicolumn{1}{l|}{\emph{GAD} \cite{guo-etal-2021-global}}  &\multicolumn{1}{l|}{BERT}    & 66.6     & 71.8     & \multicolumn{1}{c|}{69.1}     & 65.0     & 70.1     & 67.5     \\
        &\multicolumn{1}{l|}{\emph{MDCSpell} \cite{zhu2022mdcspell}}     &\multicolumn{1}{l|}{BERT}    & 70.2      & 68.8     & \multicolumn{1}{c|}{69.5}     & 69.0     & 67.7     & 68.3     \\ 
        &\multicolumn{1}{l|}{\emph{DORM} \cite{liang2023disentangled}}     &\multicolumn{1}{l|}{ChineseBERT}   & 69.5      & 73.1     & \multicolumn{1}{c|}{71.2}     & 68.4     & 71.9     & 70.1     \\ 
        &\multicolumn{1}{l|}{\emph{DR-CSC} \cite{huang2023frustratingly}}     &\multicolumn{1}{l|}{ChineseBERT}   & 70.2      & 73.3     & \multicolumn{1}{c|}{71.7}     & 69.3     & 72.3     & 70.7     \\ 
        \cline{2-9}
        &\multicolumn{1}{l|}{PLM-FT (ChineseBERT)}     &\multicolumn{1}{l|}{ChineseBERT}   & 66.7     & 70.4      & \multicolumn{1}{c|}{68.5}      & 65.0    & 68.7     & 66.8     \\ 
        &\multicolumn{1}{l|}{\textit{Bi-DCSpell} (BERT)}     &\multicolumn{1}{l|}{BERT}   & 69.9     & 70.9     & \multicolumn{1}{c|}{70.4}       & 68.5     & 68.7     & 68.6     \\ 
        &\multicolumn{1}{l|}{\textit{Bi-DCSpell} (ChineseBERT)}     &\multicolumn{1}{l|}{ChineseBERT}   & \textbf{70.9}     & 73.7     & \multicolumn{1}{c|}{\textbf{72.3}}       & \textbf{69.6}     & 72.4     & \textbf{71.0}     \\ \hline
    \multirow{12}{*}{SIGHAN15}    
        &\multicolumn{1}{l|}{\emph{Soft-Masked BERT $^{\ast}$} \cite{zhang2020spelling}}      &\multicolumn{1}{l|}{BERT}    & 67.6      & 78.7      & \multicolumn{1}{c|}{72.7}      & 63.4      & 73.9      & 68.3      \\
        &\multicolumn{1}{l|}{\emph{SpellGCN} \cite{cheng2020spellgcn}}    &\multicolumn{1}{l|}{BERT}    & 74.8    & 80.7    & \multicolumn{1}{c|}{77.7}     & 72.1    & 77.7    & 75.9    \\
        &\multicolumn{1}{l|}{\emph{MLM-phonetics} \cite{huang2021phmospell}}    &\multicolumn{1}{l|}{BERT}    & 77.5     & 83.1     & \multicolumn{1}{c|}{80.2}        & 74.9     & 80.2     & 77.5    \\
        &\multicolumn{1}{l|}{\emph{DCN} \cite{wang-etal-2021-dynamic}}     &\multicolumn{1}{l|}{BERT}    & 77.1     & 80.9     & \multicolumn{1}{c|}{79.0}     & 74.5     & 78.2     & 76.3     \\
        &\multicolumn{1}{l|}{\emph{GAD} \cite{guo-etal-2021-global}} &\multicolumn{1}{l|}{BERT}    & 75.6     & 80.4     & \multicolumn{1}{c|}{77.9}     & 73.2     & 77.8     & 75.4        \\
        &\multicolumn{1}{l|}{\emph{MDCSpell} \cite{zhu2022mdcspell}}     &\multicolumn{1}{l|}{BERT}    & 80.8      & 80.6     & \multicolumn{1}{c|}{80.7}     & 78.4     & 78.2     & 78.3     \\ 
        &\multicolumn{1}{l|}{\emph{CoSPA} \cite{yang2022cospa}}     &\multicolumn{1}{l|}{ChineseBERT}   & 79.0      & 82.4     & \multicolumn{1}{c|}{80.7}     & 76.7     & 80.0     & 78.3     \\ 
        &\multicolumn{1}{l|}{\emph{DORM} \cite{liang2023disentangled}}     &\multicolumn{1}{l|}{ChineseBERT}   & 77.9      & 84.3     & \multicolumn{1}{c|}{81.0}     & 76.6     & 82.8     & 79.6     \\ 
        &\multicolumn{1}{l|}{\emph{DR-CSC} \cite{huang2023frustratingly}}     &\multicolumn{1}{l|}{ChineseBERT}   & \textbf{82.9}      & 84.8     & \multicolumn{1}{c|}{83.8}     & \textbf{80.3}     & 82.3     & 81.3     \\ 
        \cline{2-9}
        &\multicolumn{1}{l|}{PLM-FT (ChineseBERT)}     &\multicolumn{1}{l|}{ChineseBERT}   & 79.3     & 83.0      & \multicolumn{1}{c|}{81.1}      & 77.2    & 80.7     & 78.9     \\ 
        &\multicolumn{1}{l|}{\textit{Bi-DCSpell} (BERT)}     &\multicolumn{1}{l|}{BERT}   & 79.6     & 82.4     & \multicolumn{1}{c|}{81.0}      & 77.5     & 80.2     & 78.8     \\ 
        &\multicolumn{1}{l|}{\textit{Bi-DCSpell} (ChineseBERT)}     &\multicolumn{1}{l|}{ChineseBERT}   & 82.6     & \textbf{85.4}     & \multicolumn{1}{c|}{\textbf{84.0}}      & 80.2     & \textbf{84.1}     & \textbf{82.1}     \\ \hline
    \end{tabular}
\caption{Experimental results on SIGHAN13, SIGHAN14 and SIGHAN15 test sets. Each model includes sentence-level precision, recall, and F1 score for both detection and correction. $^{\ast}$Due to the incompatibility of character-level results in the original paper~\cite{zhang2020spelling}, the results for \emph{Soft-Masked BERT} in the table are sourced from \cite{zhang2021correcting}, maintaining consistency in training data and metrics with our method. We evaluated \textit{Bi-DCSpell} using two backbones, Chinese-BERT-wwm (BERT) and ChineseBERT for a fair comparison.}
\label{results}
\end{table*}

\subsection{Model Settings}

Considering that \textit{Bi-DCSpell} is essentially a fine-tuning model for CSC, a wide range of CSC methods based on fine-tuning are selected as comparison models\footnote{Note that, due to the inherent disparity between a fine-tuned model and one trained through a combination of pre-training and fine-tuning, we do not choose the models relying on pre-training for comparison, such as the current state-of-the-art CSC model PTCSpell~\cite{wei-etal-2023-ptcspell}}: \emph{PLM-FT} is a fine-tuned PLM on the training data, which is indeed a standard \textit{C-only} model and can be viewed as a simplified version of \textit{Bi-DCSpell} without the interactive learning module and the detection encoder. 
\emph{Soft-Masked BERT} \cite{zhang2020spelling}, \emph{MLM-phonetics} \cite{zhang2021correcting} and DR-CSC \cite{huang2023frustratingly} utilize a pipeline model, they all use the information of the previous step as the prior information of the next step. 
\emph{SpellGCN} \cite{cheng2020spellgcn} incorporates phonological and visual similarity knowledge representation into BERT by employing a specialized graph convolutional network. 
\emph{DCN} \cite{wang-etal-2021-dynamic} uses a dynamically connected network to measure the degree of dependence between any two adjacent Chinese characters. 
\emph{CoSPA} \cite{yang2022cospa} reports an alterable copy mechanism to increase the generation probability of the original input, thereby mitigating the over-correction. 
\emph{DORM} \cite{liang2023disentangled} allows the direct interaction between textual and phonetic information. 
\emph{MDCSpell} \cite{zhu2022mdcspell} utilizes multi-task learning to jointly model detection and correction, implementing a multi-task learning model with task-specific feature encoders similar to those in \textit{Bi-DCSpell}. 
However, it neglects the knowledge interaction between detection and correction.

\subsection{Main Results}
The experimental results are shown in Table~\ref{results}. Note that the performance of \textit{Bi-DCSpell} depends on various hyper-parameters including $\alpha$ and $\beta$ for controlling the degree of interaction, and $\lambda$ for balancing detection and correction objectives in the loss function. The best performing results are reported in Table~\ref{results}, while the performance of different settings of these hyper-parameters are discussed in more detail in Section~\ref{Ablation_Study}.

Overall, our proposed \textit{Bi-DCSpell} consistently achieves the best F1 score for both detection and correction on all three SIGHAN datasets. This result indicates the effectiveness of our method.

In the same scenario (i.e., backbone is ChineseBERT), compared with \emph{PLM-FT}, which is a \textit{C-only} baseline, \textit{Bi-DCSpell} significantly improves the correction F1 score by 2.8\%, 4.2\%, and 3.0\% respectively, validating the necessity of modeling the interactions between detection and correction.

Compared with BERT-based \emph{Soft-Masked BERT}, \emph{MLM-phonetics}, \emph{MDCSpell}, \emph{CoSPA} and ChineseBERT-based \emph{DR-CSC}, which all adopt uni-directional interaction framework (i.e., \textit{D2C}), \textit{Bi-DCSpell} performs significantly better on all the datasets in both scenarios, confirming that modeling bi-directional interactions between detection and correction can further improve the CSC performance.

Compared with \emph{MLM-phonetics}, \emph{CoSPA} and \emph{SpellGCN} with BERT backbone, which all explicitly introduce extra phonological and/or visual information into the inference process, the proposed \textit{Bi-DCSpell} (BERT) does not incorporate any external information but it still achieves better F1 scores on almost all datasets. This result demonstrates the competitive performance and robustness of our proposed method.

Compared with \emph{DORM} that modeled the direct interaction between textual and phonetic information in the same scenario, \textit{Bi-DCSpell} (ChineseBERT) achieves better F1 scores on all datasets.

\subsection{Ablation Study}
\label{Ablation_Study}

In this subsection, we use SIGHAN15 test sets to examine the effects of various core components and parameters of our model, including the interactive learning module, the number of bi-directional interaction layers, the control gate factors $\alpha$ and $\beta$, and the balancing hyperparameter $\lambda$.

%
\begin{table}[t]
\centering
\small
\renewcommand{\arraystretch}{0.9}   
    \begin{tabular}{l|c|c|c} 
    \hline
    \textbf{Method} & \textbf{Interaction}  & \textbf{Det-F1} & \textbf{Cor-F1} \\ \hline
    PLM-FT                     & no                & 81.1   & 78.9  \\
    D2C    & uni-directional   & 82.7$_{+1.6}$   & 80.7$_{+1.8}$  \\
    \textit{Bi-DCSpell}        & bi-directional        & 84.0$_{+\bm{2.9}}$   & 81.9$_{+\bm{3.0}}$  \\ \hline
    \end{tabular}
\caption{The results of three interaction frameworks on the SIGHAN15 test set. D2C is the Bi-DCSpell with the fixed gate $\alpha=0.0$.}
\label{Ablation}
\end{table}
\textbf{Bi-directional interactive learning.}  Table~\ref{Ablation} shows a result comparison between three different frameworks, including \emph{PLM-FT} (\textit{C-only}), \emph{\textit{D2C}} and \emph{Bi-DCSpell}. 
The results illustrate that incorporating the interactions between detection and correction lead to significant performance improvements, and the modeling of bi-directional interactions brings more benefits than considering uni-directional interactions only.
\begin{figure}[t]
    \centering
    \includegraphics[width=0.4\textwidth]{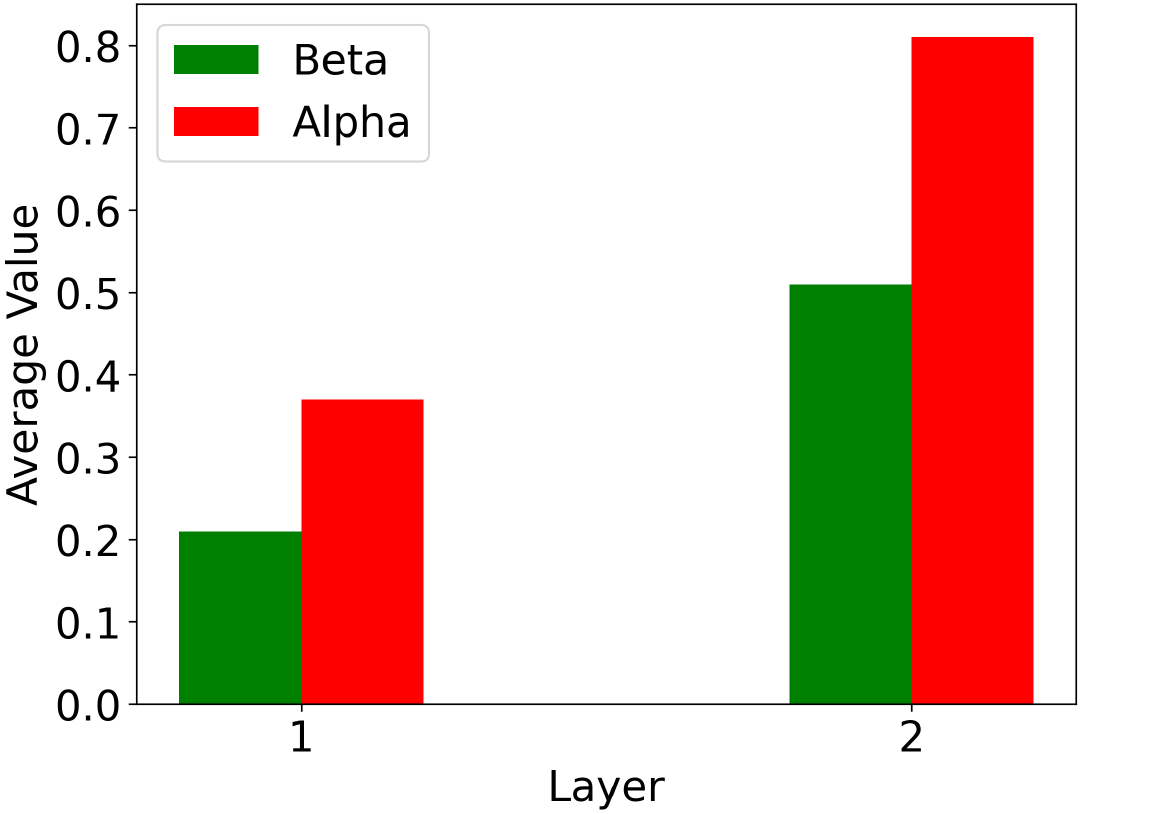}
    \caption{The average value of $\alpha$ and $\beta$ with different bi-directional interaction layers.}
    \label{gates_values}
\end{figure}

\textbf{The range of optimal degree of interaction.} The \emph{Bi-DCSpell} framework contains two learnable controlling gates: $\alpha$ and $\beta$. 
%
We compute the mean values of $\alpha$ and $\beta$ across different interaction layers in the optimal model scenario, as illustrated in Figure~\ref{gates_values}. In both layers, $\alpha$ and $\beta$ are less than 1, indicating a moderate level of bi-directional interaction. And the information exchange from correction to detection is also lower than from detection to correction. Specifically, in the first layer, both parameters reflect a low degree of interaction, while in the second layer, the bi-directional interaction becomes more pronounced.
%
%
\begin{figure}[t]
    \centering
    \includegraphics[width=0.5\textwidth]{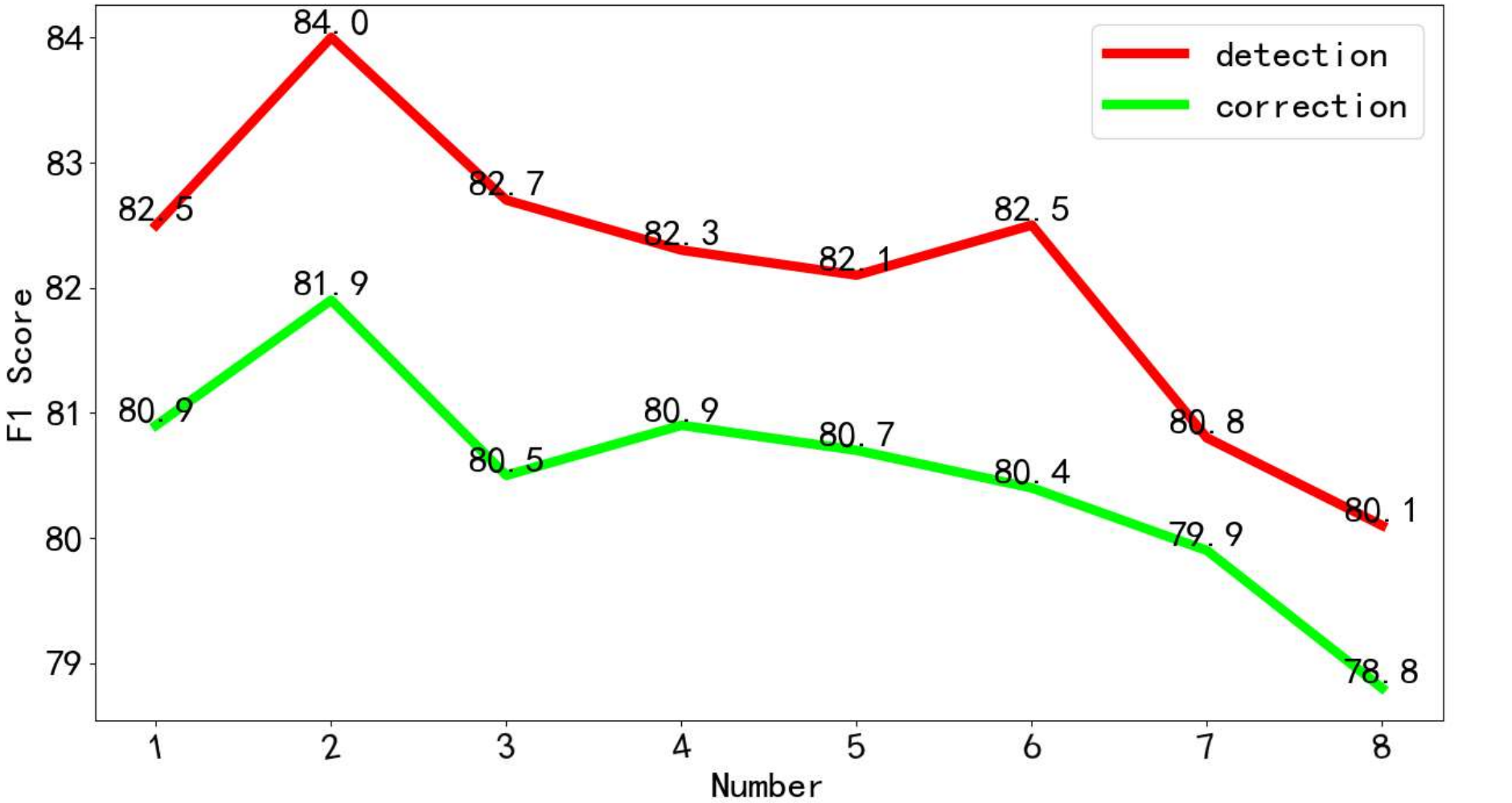}
    \caption{The F1 scores for detection and correction vary as the number of bi-directional interaction layers increases.}
    \label{numbers}
\end{figure}
%

\begin{table*}[t]
\small
\centering
\renewcommand{\arraystretch}{1.0}   
\begin{tabular}{l|l|l}
\hline
\multirow{6}{*}{case 1} 
    & Input              
        & \begin{CJK}{UTF8}{gbsn} 团长将士兵\textcolor{red}{布(bù)}署(shǔ)在城外，让他们\textcolor{red}{安(ān)}兵不动。 \end{CJK} \\ \cline{2-3} 
    & \textit{C-only}                  
        & \begin{CJK}{UTF8}{gbsn} 团长将士兵\textcolor{red}{布置(bù zhì)}在城外，让他们\textcolor{blue}{按(àn)}兵不动。\end{CJK} \\ \cline{2-3} 
    & \textit{D2C}
        & 0  0  0  0  0  \textcolor{red}{0}  0  0  0  0  0  0  0  0 \textcolor{blue}{1} 0 0 0 0 \\ 
    &                    & \begin{CJK}{UTF8}{gbsn} 团长将士兵\textcolor{red}{布(bù)}署(shǔ)在城外，让他们\textcolor{blue}{按(àn)}兵不动。\end{CJK} \\ \cline{2-3} 
    & \multirow{2}{*}{\textit{Bi-DCSpell}} 
        & 0  0  0  0  0  \textcolor{blue}{1}  0  0  0  0  0  0  0  0 \textcolor{blue}{1} 0 0 0 0 \\  
    &                    & \begin{CJK}{UTF8}{gbsn} 团长将士兵\textcolor{blue}{部署(bù shǔ)}在城外，让他们\textcolor{blue}{按(àn)}兵不动。\end{CJK} \\ \cline{2-3}
    & Translation                  
        & The captain deployed the soldiers outside the city and ordered them to stand still. \\ 
\hline
\multirow{6}{*}{case 2}     
    & Input              
        & \begin{CJK}{UTF8}{gbsn} 任何\textcolor{red}{因(yīn)}难(nán)都不能\textcolor{red}{下(xià)}倒(dǎo)有坚强意志的\textcolor{red}{对(duì)}员们。 \end{CJK} \\ \cline{2-3} 
    & \textit{C-only}                  
        & \begin{CJK}{UTF8}{gbsn} 任何\textcolor{blue}{困难(kùn nán)}都不能\textcolor{red}{击败(jī bài)}有坚强意志的\textcolor{blue}{队(duì)}员们。 \end{CJK} \\ \cline{2-3} 
    & \textit{D2C} 
        & 0  0  \textcolor{red}{1  1}  0  0  0  \textcolor{red}{0}  0  0  0  0  0  0  0  \textcolor{blue}{1} 0 0 0 \\ 
    &                    & \begin{CJK}{UTF8}{gbsn} 任何\textcolor{red}{原因(yuán yīn)}都不能\textcolor{red}{下(xià)}倒(dǎo)有坚强意志的\textcolor{blue}{队(duì)}员们。 \end{CJK} \\ \cline{2-3} 
    & \multirow{2}{*}{\textit{Bi-DCSpell}} 
        & 0  0  \textcolor{blue}{1}  0  0  0  0  \textcolor{blue}{1}  0  0  0  0  0  0  0  \textcolor{blue}{1} 0 0 0 \\  
    &                    & \begin{CJK}{UTF8}{gbsn} 任何\textcolor{blue}{困难(kùn nán)}都不能\textcolor{blue}{吓倒(xià dǎo)}有坚强意志的\textcolor{blue}{队(duì)}员们。 \end{CJK}  \\ \cline{2-3}
    & Translation                  
        & No difficulty can intimidate team members with strong determination. \\ 
\hline
\end{tabular}
\caption{Examples of CSC results from \textit{Bi-DCSpell}, in comparison with results from \textit{C-only} and \textit{D2C} baselines. \textcolor{red}{Red} and \textcolor{blue}{blue} are used to mark incorrect and correct characters, respectively.}
\label{cases}
\end{table*}

\textbf{The number of bi-directional interaction layers.} As shown in Figure~\ref{numbers}, the F1 scores of detection and correction are subject to some fluctuations when the number of bi-directional interaction layers changes between 1 and 8. These fluctuations generally increase when the number varies from 1 to 2, and then show a downward trend. Especially when the number is 2, the F1 scores reach a maximum value in both detection and correction. Therefore, considering the efficiency, we choose a layer number of 2 for the remaining experiments.

\begin{figure}[t]
    \centering
    \includegraphics[width=0.5\textwidth]{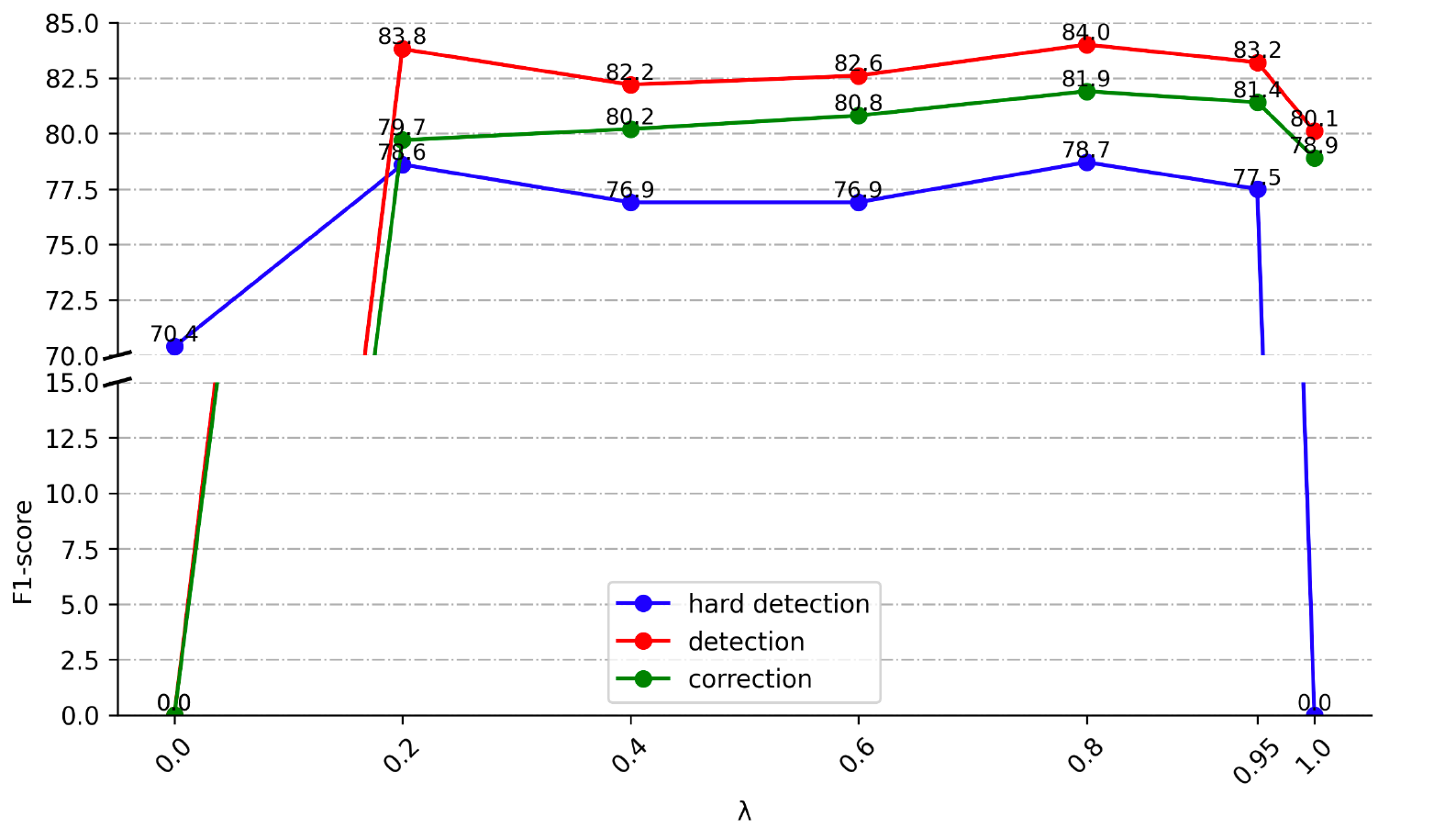}
    \caption{The F1 scores with respect to $\lambda$.
    }
    \label{lambda}
\end{figure}

\textbf{The balancing hyperparameter $\lambda$ in loss function.} Figure~\ref{lambda} provides the detection, correction and hard detection\footnote{To clarify, the evaluation of detection performance is based on the corrected sentences generated by the corrector.  The evaluation of `hard detection' is based directly on the results of the detection classifier layer.} F1 scores while  $\lambda$ increases. The result shows that the detection and correction F1 scores are $0.0$ when $\lambda$ is $0$ (where the correction classifier is not trained), and the hard detection F1 score is $0.0$ when $\lambda$ is $1$ (where the detection classifier is not trained).  Importantly, the hard detection rapidly increases when  $\lambda$ increases from $0$ to $0.2$. When the $\lambda$ goes from $0.8$ to $1.0$, the detection and correction performance are both decreasing. When the $\lambda = 0.8$, the correction F1 score reaches a maximum. Thus, we set the $\lambda$ as $0.8$ for an optimum performance of correction.  

\subsection{Case Study}
To further illustrate the effectiveness of \textit{Bi-DCSpell} in comparison with the \textit{C-only} and \textit{D2C} baselines, we analyze two concrete cases in Table~\ref{cases}. 


In the first case, the MLM-based \textit{C-only} model mis-corrected ``\begin{CJK}{UTF8}{gbsn}布署(bù shǔ)\end{CJK}" into ``\begin{CJK}{UTF8}{gbsn}布置(bù zhì)\end{CJK}", a high-frequency collocation, resulting in over-correction. On the other hand, the \textit{D2C} model did not correct ``\begin{CJK}{UTF8}{gbsn}布署(bù shǔ)\end{CJK}", due to its failure in detection. This is an example of under-correction. In contrast, the \textit{Bi-DCSpell} not only correctly detected ``\begin{CJK}{UTF8}{gbsn}布署(bù shǔ)\end{CJK}" and ``\begin{CJK}{UTF8}{gbsn}安(ān)\end{CJK}" as misspelled characters, but also made accurate corrections. 

The second case involves three discrete misspellings (``\begin{CJK}{UTF8}{gbsn}因难(yīn nán), 下倒(xià dǎo) and 对员(duì yuán)\end{CJK}"). The \textit{C-only} model correctly corrected two misspelled characters, but mis-corrected ``\begin{CJK}{UTF8}{gbsn}下倒(xià dǎo)\end{CJK}'' to a more common phrase ``\begin{CJK}{UTF8}{gbsn}击败(jī bài)\end{CJK}'' in this context, resulting in an over-correction. The \textit{D2C} model was able to address some of the issues. However, its detection module failed to identify the wrong character ``\begin{CJK}{UTF8}{gbsn}下(xià)\end{CJK}", and as a consequence its correction module was not able to correct it. Additionally, the detection module mis-identified the ``\begin{CJK}{UTF8}{gbsn}难( nán)\end{CJK}'' in ``\begin{CJK}{UTF8}{gbsn}因难(yīn nán)\end{CJK}'' as incorrect character, and in turn its correction module considered ``\begin{CJK}{UTF8}{gbsn}任何(rèn hé)\end{CJK}'' and ``\begin{CJK}{UTF8}{gbsn}原因(yuán yīn)\end{CJK}'' as a better match, leading to the mis-correction of ``\begin{CJK}{UTF8}{gbsn}因难(yīn nán)\end{CJK}'' to ``\begin{CJK}{UTF8}{gbsn}原因(yuán yīn)\end{CJK}''. On the other hand, \textit{Bi-DCSpell} successfully identified all misspelled characters and accurately corrected them, showing a better ability to tackle both over-correction and under-correction problems.

\section{Conclusions}
\label{chp5}
In this paper, we propose a novel CSC framework, namely \textit{Bi-DCSpell}, which constitutes a bi-directional interactive detector-corrector framework, mutually enhancing the feature representation for the detection and correction sub-tasks. Comprehensive experiments and empirical analyses attest to the effectiveness of our approach. The explicit modeling of bi-directional interactions between sub-tasks also holds potential significance for analogous tasks, like grammatical error correction, which deserves further research in the future.

\section{Limitations}
\subsection{Inconsistent Output}
Due to the output layer of the detector-corrector framework having two results: detection labels and correction sentences, which occupy different distribution spaces, the issue of inconsistent outputs poses a new challenge. In Bi-DCSpell, two cross-attention networks are utilized to perform feature selection from the corrector to the detector and from the detector to the corrector. During model training, the Bi-DCSpell will dynamically adjust the two tasks from joint training. And we conducted a new comparison using two consistency metrics: character-level and sentence-level. Specifically, we compare PLM-FT (ChineseBERT) and Bi-DCSpell (ChineseBERT) on the SIGHAN15 dataset. The results are as Table~\ref{consistency}:

\begin{table}[t]
\centering
\small
\renewcommand{\arraystretch}{1.0}   
    \begin{tabular}{l|c|c|c|c} 
    \hline
    \textbf{Method} & \textbf{D-F1}  & \textbf{C-F1} & \textbf{C-l} & \textbf{S-l} \\ \hline
    PLM-FT      &  81.1      & 78.9     & 95.8    & 41.9 \\
    \textit{Bi-DCSpell}   & 84.0   & 82.1   & 97.7   & 45.7  \\
    \hline
    \end{tabular}
\caption{The consistent output statistics. The `D-F1', `C-F1', `C-l' and `S-l' denote metric `detection f1-score', `correction f1-score', `character-level consistency' and `sentence-level consistency'.}
\label{consistency}
\end{table}

From the above table, compared with PLM-FT (ChineseBERT), which is a fine-tuned PLM with two classifiers for corrector and detector, respectively, Bi-DCSpell achieves better consistency scores at both the character-level and sentence-level.

\subsection{Language Limitation}
In this work, we focus only on the spelling check of Chinese characters, because CSC is very different from other languages such as English. Specifically, (1) there are no delimiters between words. (2) Chinese have more than 100,000 characters, and about 3,500 are frequently used in daily life, and most characters have similar visual and/or similar pronunciations. However, we believe that the bi-directional interactions between detection and correction is also important for spelling check in English texts, which is worth an in-depth investigation in the future.
%

\subsection{Running Efficiency}
In our code implementation, we have not paid too much attention to the running efficiency of the proposed methods. More precisely, it is expensive to take about 10 hours on an RTX A6000 (48GB GPU memory) to finish the training process. We think that the training efficiency can be improved by deploying the model training process on multiple GPUs and using data-parallel operations to increase the training batch size and shorten the training time.
%
\bibliography{custom}

\appendix

\newpage



\section{Evaluation Metrics}
\label{appendix_evaluation_metrics}
The general evaluation metrics could be divided into sentence-level and character-level metrics.
Specifically, a prediction at the sentence-level is only deemed correct if all of the misspelled characters in the sentence have been detected or corrected. Compared with character-level evaluation, sentence-level evaluation is more strict and tends to yield lower scores. Following the common practice in the literature \cite{hong2019faspell, cheng2020spellgcn}, we adopt the commonly used sentence-level precision, recall, and F1 score measures.

\section{Optimal interaction range with hyper-parameter selection}
\label{hyper-parameter selection}
\begin{figure}[t]
    \centering
    \includegraphics[width=0.5\textwidth]{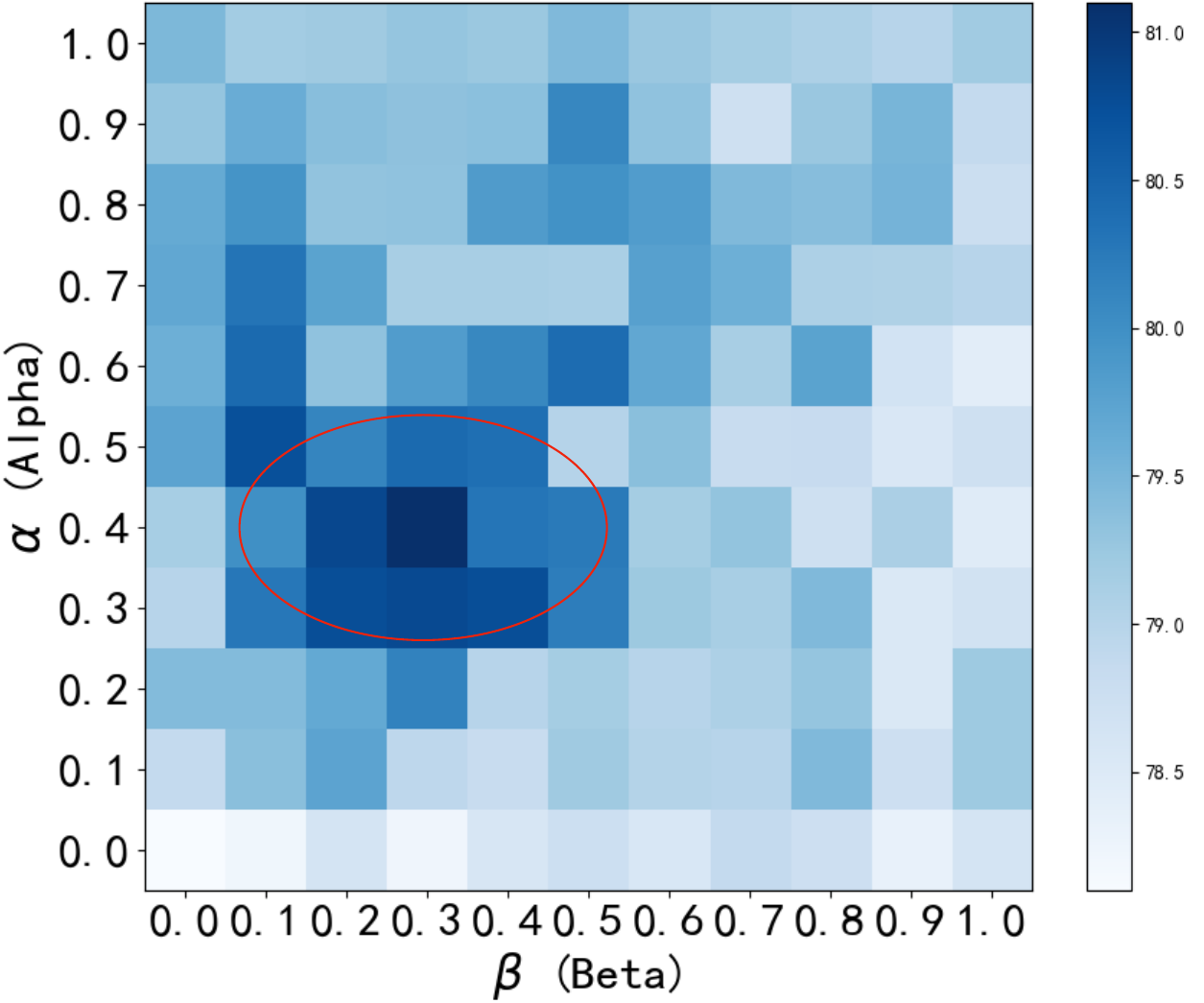}
    \caption{Correction F1 scores of Bi-DCSpell with different combinations of interaction degrees ($\alpha$ and $\beta$). The deeper color indicates the higher performance.}
    \label{gates}
\end{figure}

To verify the effectiveness of the learnable control gates $\alpha$ and $\beta$ described in Section 4.4, we further adopted a hyper-parameter selection strategy to identify their optimal ranges. We use a heatmap to visualize the impact of these two gate values (set them as parameters) on the correction F1 score, as depicted in Figure~\ref{gates}. In general, a moderate bi-directional interaction yield the highest correction F1 score, aligning with the findings in Figure~\ref{gates_values}, outperforming both unidirectional ($\alpha=0.0$ or $\beta=0.0$) and fully bi-dircetional interactions ($\alpha=1.0$ and $\beta=1.0$). Specifically, the model achieves the optimal correaction F1 score (darkest color) at ($\alpha=0.4, \beta=0.3$), with progressively lighter colors when moving away from this point. In other words, the F1 score gradually decreases, reaching a minimum at ($\alpha = 0.0$, $\beta = 0.0$) where there is no interaction.

\end{document}